# Image Recognition of Tea Leaf Diseases Based on Convolutional Neural Network

Xiaoxiao SUN[1], Shaomin MU[1], Yongyu XU[2], Zhihao CAO[1], Tingting SU[1]

*College of Information Science and Engineering, Shandong Agricultural University, Taian 271018, People's Republic of China*[1]

*College of Plant Protection, Shandong Agricultural University, Taian 271018, People's Republic of China*[2]

18264893917@163.com, msm@sdau.edu.cn, xuyy@sdau.edu.cn , czh@sdau. edu.cn, m15065806906@163.com

*Abstract*—In order to identify and prevent tea leaf diseases effectively, convolution neural network (CNN) was used to realize the image recognition of tea disease leaves. Firstly, image segmentation and data enhancement are used to preprocess the images, and then these images were input into the network for training. Secondly, to reach a higher recognition accuracy of CNN, the learning rate and iteration numbers were adjusted frequently and the dropout was added properly in the case of over-fitting. Finally, the experimental results show that the recognition accuracy of CNN is 93.75%, while the accuracy of SVM and BP neural network is 89.36% and 87.69% respectively. Therefore, the recognition algorithm based on CNN is better in classification and can improve the recognition efficiency of tea leaf diseases effectively.

## I. INTRODUCTION

Tea is one of the traditional drinks in China. The yield and quality of tea is directly related to the economic income of tea farmers and the overall economic interests of tea areas. The emergence of tea leaf diseases seriously affects the yield and quality of tea. Therefore, how to identify and control these diseases is the focus of attention in the agricultural field [1]. At present, the identification and classification of tea leaf diseases basically rely on the professional knowledge and work experience of tea farmers, then the problems of time-consuming, laborious and inefficient can seriously affect the efficiency of tea leaf disease control work. In recent years, the emergence of many machine learning algorithms provides theoretical support for the prevention of agricultural diseases, and many intelligent algorithms have become a research hotspot in the field of image recognition such as CNN and SVM.

CNN is a kind of deep feed-forward artificial neural network. Its local connection, weight sharing and pooling operation make it possible to reduce the complexity of the network effectively. It can make the model invariant to translation, distortion, scaling and other operations to a certain extent, and have strong robustness and fault tolerance. So it is easy to train and optimize the network structure [2]. Therefore, the application of CNN in the field of crop disease identification is gradually increasing. For example, in the research of Sladojevic et al. [3] proposed a kind of plant leaf disease image recognition method based on CNN, and the depth CNN model constructed has achieved the high accuracy rate to each kind of test recognition. In China, Wenxue Tan [4] proposes a method of crop disease image processing and disease recognition based on machine learning. The network structure and convolution BP error transfer algorithm are designed by using the method of disease image recognition based on elastic momentum deep convolution network, which is of positive significance for speeding up large sample machine learning process. Then a method of tobacco disease image automatic recognition based on CNN was proposed by Jing Li [5], and a Web-based tobacco disease diagnosis system was designed to identify and control tobacco diseases effectively.

However, there is no report on the identification of leaf disease image of tea leaves based on CNN. Therefore, the application of CNN in leaf disease image of tea leaves is studied in this paper. By changing the learning rate, iteration times and dropout, the network can converge to a stable value. Finally, the validity of CNN is verified and improved by comparing with other machine learning algorithms in recognition results. The recognition efficiency of tea leaf disease image has certain reference value for automatic and rapid diagnosis of tea leaf diseases.

## II. RELATED WORKS

### A. Image Acquisition

Diseased leaves of tea were carried out in tea gardens in Tai'an City, and then treated with the apparatus shown in figure 1. The picked blades are placed on the conveyor belt of white material in turn by hand. This device uses a white material to obtain a single background image, reducing background interference. In order to avoid shooting shadows and reduce image noise, the device uses LED ring light source. The position of the LED ring light source and the digital camera are fixed above the end of the conveyor belt. The camera lens takes pictures of the tea leaves on the conveyor belt from the middle hole of the LED ring light source. The model of the LED ring light source is HX7040, and the model of the digital camera is Canon EOS 80D (18-200mm). The camera is set to automatic shooting mode with an interval of 2 seconds. And conveyor belt speed is set to 10 mm / s, then a leaf diseased blade is placed on the conveyor belt every

20mm to achieve timed shooting. The image format is JPEG with 24 bitmaps, and the number of images was 15 063, including 6 kinds of common leaf diseases and mechanical damaged leaves.

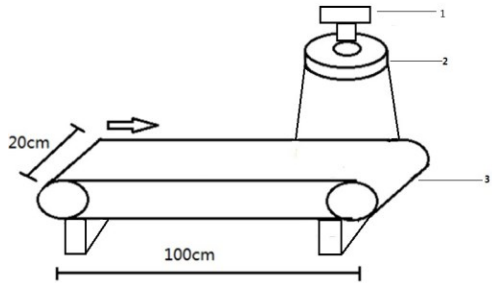

Note：1. Digital camera , 2. LED ring light source, 3 .Conveyor belt.

Fig.1 The acquisition device of experimental image

*B. Image Segmentation and Data Enhancement*

As for the different sizes of the leaves, it is necessary to locate and cut the image. In the tea leaf disease image with high definition, cutting the image can reduce the background interference information of the image, and the image target area after cutting is more obvious, which is convenient for the neural network to extract features. Therefore, this paper adopts image processing technology to realize automatic image segmentation, and its process is as follows:

(1) Rename the prepared data set for batch reading;

(2) Use Gaussian image smoothing to remove some interfering elements;

(3) Convert a color map into a grayscale image to find the image outline;

(4) Then use the Soble edge extraction method to extract the edges in the vertical direction. The formula is described as follows:

$$G=\sqrt{G_x^2 + G_y^2} \quad (1)$$

$$|G|=|G_x^2|+|G_y^2| \quad (2)$$

$$\Theta = \arctan(\frac{G_x}{G_y}) \quad (3)$$

Formula (1) shows that the gray value of a point is equal to the square root of the sum of the horizontal and vertical squares of each pixel of the point. In formula (2), in order to improve the efficiency of calculation, an approximation value without square was used as the gray value of the point. And then the direction of antimony degree is calculated by formula (3);

(5) Binary image;

(6) Use the methods of horizontal scanning and vertical scanning to locate the target area and cut it.

The comparison between before and after image segmentation of tea leaf diseases is shown in figure 2.

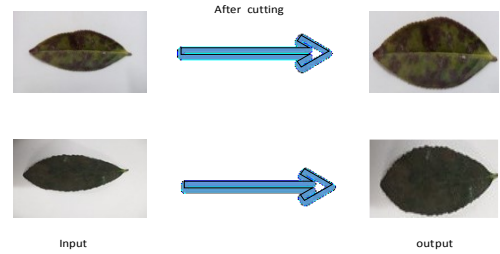

Fig.2 The comparison between before and after image segmentation

The data set image samples obtained by cutting the images are shown in table I, and the number and proportion distribution of each type of samples are shown in table II .

Table I Image samples of leaf diseases of tea leaves

| Classes | Samples |
|---|---|
| Anthracnose | |
| Leaf blight | |
| Blight disease | |
| Tea wheel spot disease | |
| Tea white star disease | |
| Tea coal disease | |
| Mechanical damage | |

Table II Number and proportion of leaf disease images in tea leaves

| Classes | Number | Proportion |
|---|---|---|
| Anthracnose | 4566 | 0.303 |
| Leaf blight | 3018 | 0.200 |
| Blight disease | 1711 | 0.114 |
| Tea wheel spot disease | 276 | 0.018 |
| Tea white star disease | 1703 | 0.113 |
| Tea coal disease | 585 | 0.039 |
| Mechanical damage | 3204 | 0.213 |

CNN is suitable for large-scale data. In order to avoid over-fitting and improve the training quality of sample set, it is necessary to increase the number of data sets appropriately [11]. Therefore, some images are enhanced before training, and more images are generated by rotating, flipping and adding noise to expand the data set. A total of 25186 images were obtained through data enhancement. The image enhancement methods we used are shown in table III.

Table III Image enhancement methods

| Methods | Specific operations |
|---|---|
| Rotate | The rotation of the original image at different angles is 90 degrees, 180 degrees and 270 degrees respectively. |
| Flipping | The original image is turned upside down. |
| Noise addition | Salt and pepper noise is added to the original image, and some pixel values can be changed randomly. On the two value image, some pixels are whitened and some pixels turn black. |

The processing effect of each image enhancement method and the number and proportion of each class of samples in the enhanced data set are shown in table IV and table V.

Table IV Enhancement methods and example diagrams

| Enhancement methods | Example diagrams | | | |
|---|---|---|---|---|
| Original image | 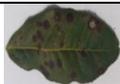 | 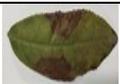 | 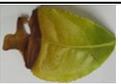 | 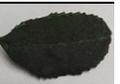 |
| Rotation | 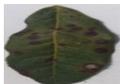 | 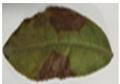 | 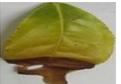 | 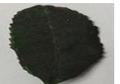 |
| Upside down | 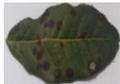 | 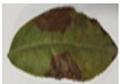 | 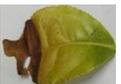 | 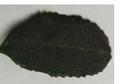 |
| Noise addition | 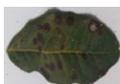 | 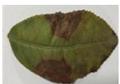 | 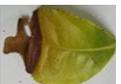 | 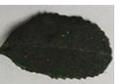 |

Table V Number and proportion of sample images after data enhancement

| Classes | Number | Proportion |
|---|---|---|
| Anthracnose | 7076 | 0.252 |
| Leaf blight | 4482 | 0.160 |
| Blight disease | 4022 | 0.143 |
| Tea wheel spot disease | 1834 | 0.065 |
| Tea white star disease | 3806 | 0.136 |
| Tea coal disease | 2038 | 0.073 |
| Mechanical damage | 4786 | 0.171 |

The sample needs to uniform size and format, so all images are scaled to 188×188, and the training data set and the test data set are established in a ratio close to 9:1 in all disease images. Then 25186 images are used as training data set, and the rest 2858 images are used as test sets.

### III. CNN

#### A. Overview

CNN is the most widely used deep learning model in the field of computer vision [6]. It was first proposed by Fukushima, a Japanese scholar, to imitate the mechanism of biological vision system. As a multi-layer and incompletely connected neural network, it can input the original image directly to the neural network. And it mainly including feature extraction and classification [7]. The feature extraction module is composed of convolution layer and subsampling layer alternately. The convolution layer analyzes each small block of neural network more deeply by convolution filtering to obtain more abstract useful features. The subsampling layer can further reduce the number of nodes in the last fully connected layer by sampling and reducing the data of the convolutional layer, which can reduce the parameters in the neural network. Classification module uses classifier to recognize and classify the extracted features. Classifier usually uses one or two layers of fully connected neural network to achieve. CNN can identify the changing models, and the robustness of the geometric changes is good.

#### B. Activation Function

Activation function is very important for neural network model to learn and understand very complex and nonlinear functions, which can help the network to better fit. It can introduce nonlinear characteristics into the network. Its main purpose is to transform the input signal of a node into an output signal in the CNN model.

At present, the commonly used activation functions are sigmoid, tanh and ReLU, The former two are more common used in the full-connection layer, and the latter in the convolution layer [8]. In this paper, the activation functions of the convolutional layer and the fully connected layer are both ReLU. Since ReLU is a linear piecewise function, its pre-transmission, post-transmission and derivation are piecewise linear, so it is easier to learn and optimize. Therefore, the use of ReLU allows the network to introduce sparsity on its own, while greatly improving the training speed.

*C. CNN Network Structure*

The CNN model used in the experiment has 9 layers, which contains 2 convolution layers, 2 maximum pooling layers, 2 partial response normalization layers, 2 full connection layers and 1 classification output layer. In experiments, to reduce over-fitting, then a dropout mechanism is introduced after the fully connected layer. The model diagram is shown in figure 3.

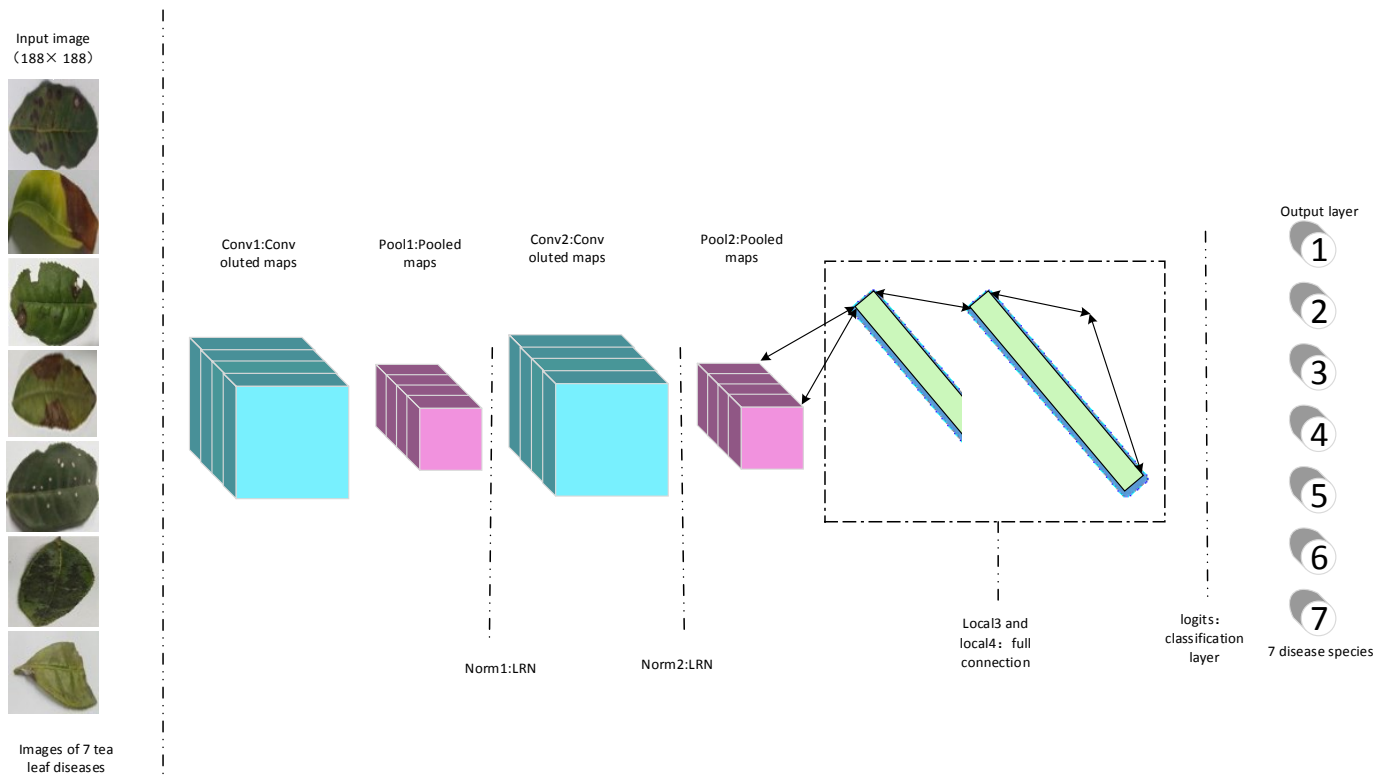

Fig.3 The structure of convolutional neural network model

(1) Convolution layer

The convolution layer is composed of several feature maps. Each feature map is composed of several neurons. Each neuron of the convolution layer is connected with the local area of the upper feature map through the convolution kernel. It can maximally extract the features of the original signal, thus enhancing the original signal features and reducing noise interference. In this paper, the gray scale pixel matrix of 188×188 tea leaf disease image is used as input. The convolution kernel size is 3×3, and the number of convolution kernels is 16, which corresponds to 16 feature maps. The size of each feature map is (188-3+1)×(188-3+1)=186×186, the number of required parameters is (3×3+1)×16=160, and the number of connections is 160×186×186=5535360.

(2) Pooling layer

Since too many features are easy to cause over-fitting, the maximum pooling algorithm (Max-pooling) is used to aggregate features from different locations and calculate the maximum value of a feature in the selected region of the image as the eigenvalue. This layer selects a maximum pool layer with a size of 3 x 3 and a step size of 2 x 2 to process the data. The maximum pool size here is not consistent with the step size, which can increase the richness of the data.

(3) Local response normalization layer

In order to improve the generalization ability of the model, the Local Response Normalization (LRN) [9] layer is introduced. LRN was first seen in papers that participated in the ImageNet competition with CNN. In addition, LRN can select larger feedback from multiple convolution kernels, especially for ReLU, an activation function with no upper bound. Therefore, using LRN to catalyze the peak value of the data and suppress the surroundings can make the data more obvious and improve the recognition efficiency.

(4) Full connection layer and output layer

The model contains two fully connected layers, which can connect all the features and send the output value to the classifier. In the process of experiment, if the network has an over-fitting phenomenon, the dropout layer can be connected in the latter full-connection layer. Weight attenuation is realized by modifying the cost function while the dropout is realized by modifying the neural network structure. And it is a small optimization method used in training the neural network. The output of the dropout layer is connected to the output layer. The output layer uses the softmax regression classifier for solving multi-classification problems. It can avoid over-fitting caused by local convergence [10].

## IV. EXPERIMENTS

### A. Experimental Environment

Anaconda is a Python distribution for scientific computing that supports Linux, Mac and Windows systems. It can provide the function of package management and environment management. Tensorflow is not only an interface to realize machine learning, but also a framework for executing machine learning algorithms. For large-scale neural network training, it allows users to implement parallel computing easily. This article installs the Tensorflow framework under Anaconda, which is easy to debug with Spyder.

### B. Comparison of Recognition Accuracy under Different Learning Rates and Steps

CNN model is used to automatically learn the image features of tea leaf diseases and classify them. Since deep learning technology combines image feature extraction with classifier training, the recognition accuracy is greatly improved. In this paper, the initial iteration number is 50 000 times, the step size is 100, and the learning rate is 0.0001. The value of learning rate determines the step size of gradient descent in back-propagation. If the value is too small, it will not achieve faster convergence, otherwise it will easily fall into local extreme value and make the training fail. The experiments were conducted with different learning rates.

Table VI Experimental accuracy under different learning rates

| Learning rate | 0.0001 | 0.0001 | 0.0001 | 0.0001 | 0.00002 | 0.00003 | 0.00004 |
|---|---|---|---|---|---|---|---|
| Number of iterations | 50000 | 50000 | 40000 | 40000 | 40000 | 40000 | 40000 |
| Dropout | | √ | | √ | | | |
| Accuracy | Over-fitting | 87.5% | over-fitting | under-fitting | under-fitting | under-fitting | under-fitting |
| Learning rate | 0.00004 | 0.00005 | 0.00006 | 0.00007 | 0.00008 | 0.00009 | 0.00009 |
| Number of iterations | 50000 | 40000 | 40000 | 40000 | 40000 | 40000 | 40000 |
| Dropout | | | | | | | √ |
| Accuracy | 93.75% | 93.75% | 93.75% | 93.75% | 93.75% | over-fitting | 87.50% |

As table VI shows, when the number of iteration steps is 50000 and the learning rate is 0.0001, a phenomenon of over-fitting occurs. Therefore, the dropout is added to the experiment, and finally the network converges to a stable value of 87.50%, but the accuracy is not high. When the number of iteration steps is 40 000, the learning rate converges to a stable value of 93.75% in the range of 0.00004 to 0.00008 without dropout, and the other conditions appear over-fitting and under-fitting respectively. Under the condition of under-fitting, the number of iterations can be increased appropriately or the learning rate can be adjusted in the experiment. For example, when the number of iteration steps is 40 000 and the learning rate is 0.00004, the number of iteration steps is increased to 50 000 to cope with the phenomenon of under-fitting, which makes the network converge steadily. In the case of over-fitting, a dropout layer is added to network training. For example, dropout is added to cope with the over-fitting phenomenon when the iteration step is 40000 and the learning rate is 0.00009, but the experimental accuracy is 87.50%.

In order to analyze the optimal learning rate, it is necessary to observe the number of iterative steps used when the same experimental accuracy is achieved. figure 4 gives the curves of experimental accuracy corresponding to normal convergence under different conditions.

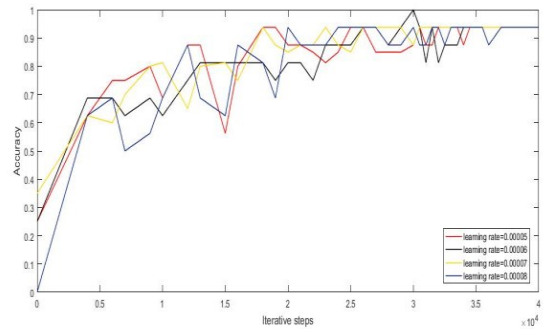

Fig.4 Comparison of experimental accuracy under different conditions

When the learning rate equals 0.00005, 0.00006, 0.00007 and 0.00008, the network converges to a stable value of 93.75%. When the number of iterative steps reaches 30500, the curve at the learning rate equal to 0.00007 converges first, then the experimental result is better, and the curve corresponding to other learning rates converges later. At 34000 steps, it basically tends to a stable value and no longer changes. Only when the learning rate is equal to 0.00008, there is a short-term instability and eventually convergence. In summary, when the learning rate is equal to 0.00007, the experimental results are good and the experimental accuracy is 93.75%.

*C. Comparison of Recognition Accuracy between CNN and Other Machine Learning Algorithms*

The experiment is mainly compared with SVM and BP algorithm, and the comparison result is shown in table VII.

Table VII Comparison of recognition accuracy between CNN and SVM and BP algorithms

| Model | Accuracy |
|---|---|
| CNN | 93.75% |
| SVM | 89.36% |
| BP | 87.69% |

As for the deep learning technology combines feature extraction with classifier training, the recognition accuracy and learning efficiency of tea diseases are greatly enhanced. It should be pointed out that the two algorithms of SVM and BP need a series of methods to pre-process the image before classification of the image of tea leaf diseases, and then extract the color, shape and texture features of the pre-processed image of tea leaf diseases. Then these feature selection classifiers are used for recognition and classification. However, the extraction of features and the complex process of parameter adjustment will increase the experimental time, thus affecting the recognition results to a certain extent. CNN, unlike SVM and BP, can input the original image directly into the network without the preprocessing process of feature extraction, greatly saving time and reducing the limitations of artificial design features. The experimental results show that the recognition accuracy of CNN is 93.75% compared with other machine learning algorithms, which verifies the effectiveness of CNN algorithm.

## V. CONCLUSION

In this paper, CNN model is used to recognize the image of tea leaf diseases. ReLU linear function is used as the activation function of neurons to accelerate the convergence of the network. By adjusting the learning rate, the number of iterations and the addition of dropout, the network can achieve high recognition accuracy, which is conducive to the identification and classification of tea diseases. In the process of experiment, the adjustment of learning rate will consume a lot of time, so it is necessary to develop a method to find the optimal learning rate, which is worthy of further study. Due to the different time and degree of tea disease occurrence, resulting in the imbalance of the number of experimental data set. So it is necessary to respond to the imbalance of data set.

ACKNOWLEDGMENTS

This work is supported by National Natural Science Foundation of Shandong Province, China (ZR2018QF002), and is supported by The Modern Tea Industry Technology System of Shandong Province，China (SDAIT-19-04). And this work was also supported by First Class Discipline Funding of Shandong Agricultural University.